\documentclass{article} 
\usepackage{iclr2019_conference}
\usepackage{natbib}
\usepackage{enumitem,amssymb}
\usepackage{amsmath}
\usepackage{amsfonts}
\usepackage{marginnote}
\usepackage{graphicx}
\usepackage[T1]{fontenc}
\usepackage{ifthen}
\usepackage{url}
\usepackage{fancyhdr}
\usepackage{hyperref}

\iclrfinalcopy 
\begin{document}

\title{\vspace{5em}On Evaluating Adversarial Robustness}
\fancyhead[R]{On Evaluating Adversarial Robustness}
\thispagestyle{empty}
\maketitle

  Nicholas Carlini\textsuperscript{1},
  Anish Athalye\textsuperscript{2},
  Nicolas Papernot\textsuperscript{1},
  Wieland Brendel\textsuperscript{3},
  Jonas Rauber\textsuperscript{3},
  Dimitris Tsipras\textsuperscript{2},
  Ian Goodfellow\textsuperscript{1},
  Aleksander M\k{a}dry\textsuperscript{2},
  Alexey Kurakin\textsuperscript{1}\,\textsuperscript{*}
  \\
  \noindent \\
  \noindent \\
  \textsuperscript{1} Google Brain
  \textsuperscript{2} MIT 
  \textsuperscript{3} University of T\"ubingen
  \\
  \noindent \\
  \noindent \\
  \textsuperscript{*} List of authors is dynamic and subject to change.
  Authors are ordered according to the amount of their contribution
  to the text of the paper.

\vspace{20em}
Please direct correspondence to the GitHub repository \\
\url{https://github.com/evaluating-adversarial-robustness/adv-eval-paper} \\
\noindent \\
Last Update: 18 February, 2019. \\ 
\newpage

\reversemarginpar

  \vspace{-1em}
\begin{abstract}
    Correctly evaluating defenses against adversarial examples has proven
to be extremely difficult.
Despite the significant amount of recent work attempting
to design defenses that withstand adaptive attacks, few have
succeeded; most papers that propose
defenses are quickly shown to be incorrect.

We believe a large contributing factor is the difficulty of performing
security evaluations.
In this paper, we discuss the methodological foundations,
review commonly accepted best
practices, and suggest new methods for evaluating defenses to
adversarial examples.
We hope that both researchers developing defenses
as well as readers and reviewers who wish to understand the
completeness of an evaluation
consider our advice in order to avoid common pitfalls.

\end{abstract}

\section{Introduction}

Adversarial examples \citep{szegedy2013intriguing,biggio2013evasion},
inputs that are specifically designed by an
adversary to force a machine learning system to produce erroneous outputs, 
have seen significant study in
recent years.
This long line of research
\citep{dalvi2004adversarial,lowd2005adversarial,barreno2006can,barreno2010security,globerson2006nightmare,kolcz2009feature,barreno2010security,biggio2010multiple,vsrndic2013detection}
has recently begun seeing significant study as machine learning
becomes more widely used.
While attack research (the study of
adversarial examples on new domains or under new threat
models) has flourished,
progress on defense%
\footnote{This paper uses the word ``defense'' with the
understanding that there are non-security
motivations for constructing machine learning
algorithms that are robust to attacks
(see Section~\ref{sec:defense_research_motivation});
we use this consistent terminology for simplicity.}
research (i.e., building systems that
are robust to adversarial examples) has been comparatively slow.

More concerning than the fact that progress is slow is the fact that
most proposed defenses are quickly shown to have performed
incorrect or incomplete evaluations
\citep{carlini2016defensive,carlini2017towards,brendel2017comment,carlini2017adversarial,he2017adversarial,carlini2017magnet,athalye2018obfuscated,engstrom2018evaluating,athalye2018robustness,uesato2018adversarial,mosbach2018logit,he2018decision,sharma2018bypassing,lu2018limitation,lu2018blimitation,cornelius2019efficacy,carlini2019ami}.
As a result, navigating the field and identifying genuine progress becomes particularly hard.

Informed by these recent results, this paper provides practical advice
for evaluating defenses that are intended to be robust to adversarial examples.
This paper is split roughly in two:
\begin{itemize}
\item \S\ref{sec:doing_good_science}: \emph{Principles for performing defense evaluations.}
  We begin with a discussion of the basic principles and methodologies that should guide defense evaluations.
\item \S\ref{sec:dont_do_bad_science}--\S\ref{sec:analysis}: \emph{A specific checklist for avoiding common evaluation pitfalls.}
  We have seen evaluations fail for many reasons; this checklist outlines
  the most common errors we have seen in defense evaluations so they can be
  avoided.
\end{itemize}
We hope this advice will be useful to both
those building defenses (by proposing evaluation methodology and
suggesting experiments that should be run)
as well as readers or reviewers of defense papers (to identify potential
oversights in a paper's evaluation).

We intend for this to be a living document.
The LaTeX source for the paper is available at
\url{https://github.com/evaluating-adversarial-robustness/adv-eval-paper} and we encourage researchers to participate
and further improve this paper.

\newpage

\section{Principles of Rigorous Evaluations}
\label{sec:doing_good_science}

\subsection{Defense Research Motivation}
\label{sec:defense_research_motivation}

Before we begin discussing our recommendations for performing defense evaluations,
it is useful to briefly consider \emph{why} we are performing the evaluation in
the first place.
While there are many valid reasons to study defenses to adversarial examples, below are the
three common reasons why one might be interested in evaluating the
robustness of a machine learning model.

\begin{itemize}

\item \textbf{To defend against an adversary who will attack the system.}
  Adversarial examples are a security concern.
  Just like any new technology not designed with security in mind,
  when deploying a machine learning system in the real-world,
  there will be adversaries who wish to cause harm as long as there
  exist incentives (i.e., they benefit from the system misbehaving).
  Exactly what this
  harm is and how the adversary will go about causing it depends on the details of
  the domain and the adversary considered.
  For example, an attacker may wish to cause a self-driving car to
  incorrectly recognize road signs\footnote{While this threat model is often repeated
  	in the literature, it may have limited impact for
  	real-world adversaries, who in practice may have
        have little financial motivation to
  	cause harm to self-driving cars.}~\citep{papernot2016limitations},
  cause an NSFW detector to incorrectly
  recognize an image as safe-for-work~\citep{bhagoji2018practical},
  cause a malware (or spam) classifier
  to identify a malicious file (or spam email) as benign~\citep{dahl2013large}, cause an
  ad-blocker to incorrectly identify an advertisement as natural content~\citep{tramer2018ad},
  or cause a digital assistant to incorrectly recognize commands it is given \citep{carlini2016hidden}.

\item \textbf{To test the worst-case robustness of machine learning algorithms.}
  Many real-world environments have inherent randomness that is difficult to
  predict.
  By analyzing the robustness of a model from the perspective of an
  adversary, we can estimate the \emph{worst-case} robustness
  in a real-world setting.
  Through random testing,
  it can be difficult to distinguish a system that fails
  one time in a billion from a system that never fails: even when evaluating
  such a system on a million choices of randomness, there is just under $0.1\%$ chance
  to detect a failure case.

  However, analyzing the worst-case robustness can discover a difference.
  If a powerful adversary who is intentionally trying to cause a system to misbehave (according
  to some definition) cannot
  succeed, then we have strong evidence that the system will not misbehave due to
  \emph{any} unforeseen randomness.

\item \textbf{To measure progress of machine learning algorithms
  towards human-level abilities.}
  To advance machine learning algorithms it is important to understand where they fail.
  In terms of performance,
  the gap between humans and machines is quite small on many widely
  studied problem domains, including reinforcement
  learning (e.g., Go and Chess \citep{silver2016mastering})
  or natural image classification \citep{krizhevsky2012imagenet}.
  In terms of adversarial robustness, however, the gap
  between humans and machines is astonishingly large: even in
  settings where machine learning achieves
  super-human accuracy, an adversary can often introduce perturbations that
  reduce their accuracy to levels of random guessing and far below the
  accuracy of even the most uninformed human.%
  \footnote{Note that time-limited humans appear
    vulnerable to some forms of adversarial examples~\citep{elsayed2018adversarial}.}
  This suggests a fundamental difference of the
  decision-making process of humans and machines.
  From this point of
  view, adversarial robustness is a measure of progress in machine
  learning that is orthogonal to performance.

\end{itemize}

The motivation for why the research was conducted informs the
methodology through which it should be evaluated:
a paper that sets out to prevent a real-world adversary from fooling a
specific spam detector assuming the adversary can not directly access
the underlying model will have a very different evaluation than one that
sets out to measure the worst-case robustness of a self-driving car's vision system.

This paper therefore does not (and could not) set out to provide a definitive
answer for how all evaluations should be performed.
Rather, we discuss methodology that we believe is common to most evaluations.
Whenever we provide recommendations that may not apply to some class of
evaluations, we state this fact explicitly.
Similarly, for
advice we believe holds true universally, we discuss why this is the case,
especially when it may not be obvious at first.

The remainder of this section provides an overview of the basic methodology
for a defense evaluation.

\subsection{Threat Models}
\label{sec:threatmodel}
A threat model specifies the conditions under which a defense
is designed to be secure and the precise security guarantees provided;
it is an integral component of the defense itself.

Why is it important to have a threat model? In the context of a defense where
the purpose is motivated by security, the threat model outlines what type of
actual attacker the defense intends to defend against, guiding the evaluation of the defense.

However, even in the context of a defense motivated by reasons beyond
security, a threat model is necessary for evaluating the performance of the defense.
One of the defining
properties of scientific research is that it is \emph{falsifiable}:
there must exist an experiment that can contradict its claims.
Without a threat model, defense proposals are often either not falsifiable
or trivially falsifiable.

Typically, a threat model includes a set of assumptions about the adversary's \textit{goals},
\textit{knowledge}, and  \textit{capabilities}.
Next, we briefly describe each.

\subsubsection{Adversary goals}
How should we define an \textbf{adversarial example}?
At a high level, adversarial examples can be defined as inputs
specifically designed to force a machine learning system to
produce erroneous outputs.
However, the precise goal of an adversary can vary significantly
across different settings.

For example, in some cases the adversary's goal may be to simply
cause misclassification---any input being misclassified
represents a successful attack.
Alternatively,
the adversary may be interested in having the model misclassify certain
examples from a \textit{source} class into a \textit{target} class of their choice.
This has been referred to a \textit{source-target} misclassification attack~\citep{papernot2016limitations}
or \textit{targeted} attack~\citep{carlini2017towards}.

In other settings, only specific types of misclassification may be interesting.
In the space of malware detection, defenders
may only care about the specific source-target class pair where an adversary
causes a malicious program to be misclassified as benign; causing a benign program
to be misclassified as malware may be uninteresting.

\subsubsection{Adversarial capabilities}
In order to build meaningful defenses, we need to impose reasonable constraints to the attacker.
An unconstrained attacker who wished to cause harm may, for example,
cause bit-flips on the
weights of the neural network, cause errors in the data processing pipeline,
backdoor the machine learning model, or (perhaps more relevant) introduce
large perturbations to an image that would alter its semantics.
Since such attacks are outside the scope of defenses adversarial examples,
restricting the adversary is necessary for designing defenses that are
not trivially bypassed by unconstrained adversaries.

To date, most defenses to adversarial examples typically restrict the adversary to
making ``small'' changes to inputs from the data-generating
distribution (e.g. inputs from the test set).
Formally, for some natural input $x$ and similarity metric
$\mathcal{D}$, $x'$ is considered a valid adversarial example if
$\mathcal{D}(x, x')\leq \epsilon$ for some small $\epsilon$ and $x'$ is
misclassified\footnote{
  It is often required that the original input $x$ is classified
  correctly, but this requirement can vary across papers.
  Some papers consider $x'$ an adversarial example as long as it
  is classified \emph{differently} from $x$.}.
This definition is motivated by the assumption that small changes under
the metric $\mathcal{D}$ do not change the true class of the input and
thus should not cause the classifier to predict an erroneous class.

A common choice for $\mathcal{D}$, especially for the case of image
classification, is defining it as the $\ell_p$-norm between two
inputs for some $p$.
(For instance, an $\ell_\infty$-norm constraint of $\epsilon$ for image
classification implies that the adversary cannot modify any individual
pixel by more than $\epsilon$.)
However, a suitable choice of $\mathcal{D}$ and $\epsilon$
may vary significantly based on the particular task.
For example, for a task with binary features one may wish to
study $\ell_0$-bounded adversarial examples more closely
than $\ell_\infty$-bounded ones.
Moreover, restricting adversarial perturbations to be small may not
always be important: in the case of malware detection, what is
required is that the adversarial program preserves the malware
behavior while evading ML detection.

Nevertheless, such a rigorous and precise definition of the
adversary's capability, leads to well-defined measures of
adversarial robustness that are, in principle, computable.
For example, given a model $f(\cdot)$, one common way to define
robustness is the worst-case loss $L$ for a given perturbation budget,
\[
\mathop{\mathbb{E}}_{(x,y) \sim \mathcal{X}}\bigg[
  \max_{x' : \mathcal{D}(x,x') < \epsilon} L\big(f(x'),y\big) \bigg].
\]
Another commonly adopted definition is the average (or median)
minimum-distance of the adversarial perturbation,
\[
\mathop{\mathbb{E}}_{(x,y) \sim \mathcal{X}}\bigg[
  \min_{x' \in A_{x,y}} \mathcal{D}(x,x') \bigg],
\]
where $A_{x,y}$ depends on the definition of \emph{adversarial example},
e.g. $A_{x,y} = \{x' \mid f(x') \ne y\}$ for misclassification
or $A_{x,y} = \{x \mid f(x') = t\}$ for some target class $t$.

A key challenge of security evaluations is that while this
\emph{adversarial risk}~\citep{madry2017towards,uesato2018adversarial}
is often computable in theory (e.g. with optimal attacks or brute force enumeration of the considered perturbations),
it is usually intractable to compute exactly, and therefore
 in practice we must approximate this quantity.
This difficulty is at the heart of why evaluating worst-case robustness is difficult:
while evaluating average-case robustness is often as simple as sampling a few
hundred (or thousand) times from the distribution and computing the mean, such
an approach is not possible for worst-case robustness.

Finally, a common, often implicit, assumption in adversarial example
research is that the adversary has direct access to the model's input
features: e.g., in the image domain, the adversary directly manipulates the image pixels.
However, in certain domains, such as malware detection or language
modeling, these features can be difficult to reverse-engineer.
As a result, different assumptions on the capabilities of the
adversary can significantly impact the evaluation of a defense's effectiveness.

\paragraph{Comment on $\ell_p$-norm-constrained threat models.}
A large body of work studies a threat model where the adversary is 
constrained to $\ell_p$-bounded perturbations.
This threat model is highly limited and does not perfectly match real-world
threats~\citep{engstrom2017rotation,gilmer2018motivating}.
However, the well-defined nature of this threat model is helpful
for performing principled work towards building strong defenses.
While $\ell_p$-robustness does not imply robustness in more realistic
threat models, it is almost certainly the case that lack of robustness
against $\ell_p$-bounded perturbation will imply lack of robustness in more
realistic threat models.
Thus, working towards solving robustness
for these well-defined $\ell_p$-bounded threat models
is a useful exercise.

\subsubsection{Adversary knowledge.}
A threat model clearly describes what knowledge the adversary
is assumed to have.
Typically, works assume either white-box  (complete knowledge
of the model and its parameters) or black-box access (no knowledge of the model)
with varying degrees of black-box access
(e.g., a limited number of queries to the model, access to the
predicted probabilities or just the predicted class,
or access to the training data).

In general, the guiding principle of a defense's threat model
is to assume that the adversary has complete 
knowledge of the inner workings of the
defense.
It is not reasonable to assume the defense
algorithm can be held secret, even in black-box threat models.
This widely-held principle is known in the field of security as Kerckhoffs'
principle~\citep{kerckhoffs1883cryptographic}, and the opposite is known as
``security through obscurity''.
The open design of security mechanisms is
a cornerstone of the field of cryptography~\citep{saltzer1975protection}.
This paper discusses only how to perform white-box evaluations, which implies
robustness to black-box adversaries,
but not the other way around. 

\paragraph{Holding Data Secret.}
While it can be acceptable to hold some limited amount of information
secret, the defining characteristic of a white-box evaluation (as
we discuss in this paper) is that the threat model assumes
the attacker has \textbf{full knowledge} of the underlying system.

That does not mean that all information has to be available to the
adversary---it can be acceptable for the defender to hold a small
amount of information secret.
The field of cryptography, for example, is built around the idea that
one \emph{can} keep secret
the encryption keys, but the underlying algorithm is be assumed to be public.

A defense that holds values secret should justify that it is reasonable to do so.
In particular, secret information generally
satisfies at least the following two properties:
\begin{enumerate}
\item \emph{The secret must be easily replaceable.} 
  That is, there should be an efficient algorithm to generate a new secret
  if the prior one happened to be leaked.
\item \emph{The secret must be nonextractable.} An adversary who is
  allowed to query the system should not be able to extract any information
  about the secret.
\end{enumerate}

For example, a defense that includes randomness (chosen fresh) at inference
time is using secret information not available to the adversary.
As long as the distribution is known, this follows Kerckhoffs' principle.
On the other hand, if a single fixed random vector was added to the output
of the neural network after classifying an input, this would not be a good
candidate for a secret.
By subtracting the observed output of the model with
the expected output, the secret can be easily determined.

\subsection{Restrict Attacks to the Defense's Threat Model}
Attack work should always evaluate defenses under the
threat model the defense states.
For example, if a defense paper explicitly states
``we intend to be robust to $L_2$ attacks of norm no greater than
1.5'', an attack paper must restrict its demonstration of vulnerabilities
in the defense to the generation of adversarial
examples with $L_2$ norm less than 1.5. Showing something different,
e.g., adversarial examples with $L_\infty$ norm less than $0.1$,
is important and useful research%
\footnote{See for example the work of \cite{sharma2017breaking,song2018generative}
  who explicitly step outside of the threat model of the original defenses
  to evaluate their robustness.} (because it teaches the research community
something that was not previously known, namely, that this system may have
limited utility in practice), but is not a
\emph{break} of the defense: the defense never claimed to be robust to
this type of attack.

\subsection{Skepticism of Results}
When performing scientific research one must be skeptical of
all results.
As Feynman concisely put it, ``the first principle [of research] is that you
must not fool yourself---and you are the easiest person to fool.''
This is never more true than when considering security evaluations.
After spending significant effort to try and develop a defense
that is robust against attacks, it is easy to assume that the
defense is indeed robust, especially when baseline attacks
fail to break the defense.
However, at this time the authors need to completely switch
their frame of mind and try as hard as possible to show their
proposed defense is ineffective.%
\footnote{One of the reasons it is so easy to accidentally fool oneself in security
  is that mistakes are very difficult to catch. Very often attacks only fail
  because of a (correctable) error in how they are being applied. It has to be the
  objective of the defense researcher to ensure that, when attacks fail, it is
  because the defense is correct, and not because of an error in applying
  the attacks.}

Adversarial robustness is a negative goal -- for a
defense to be truly effective, one needs to show that \emph{no attack} can bypass it.
It is only that by failing to show the defense is ineffective to
adaptive attacks (see below) that we can
believe it will withstand future attack by a motivated adversary (or,
depending on the motivation of the research, that the claimed lower bound is
in fact an actual lower bound).

\subsection{Adaptive Adversaries}
\label{sec:adaptive}

After a specific threat model has been defined, the remainder of the evaluation
focuses on \emph{adaptive adversaries}\footnote{We use the word ``adaptive
  adversary'' (and ``adaptive attack'') to refer to the general notion in
  security of an adversary (or attack, respectively)
  that \emph{adapts} to what the defender has done \citep{herley2017sok,carlini2017adversarial}.}
which are adapted to the specific details of the defense and attempt to invalidate
the robustness claims that are made.

This evaluation is the most important section of any paper that develops a
defense.
After the defense has been defined, ask: \emph{what attack could possibly defeat this
  defense?} All attacks that might work must be shown to be ineffective.
An evaluation that
does not attempt to do this is fundamentally flawed.

Just applying existing adversarial attacks with default hyperparameters
is not sufficient, even if these attacks
are state-of-the-art: all existing attacks and hyperparameters
have been adapted to and tested only against 
\emph{existing} defenses, and there is a good chance these attacks
will work sub-optimally or even fail against a new defense.
A typical example is gradient masking~\citep{tramer2017ensemble},
in which defenses manipulate the model's gradients and thus prevent
gradient-based attacks from succeeding.
However, an adversary aware of
the defense may recover these gradients through a black-box input-label queries, as
shown by~\citet{papernot2017practical}, or through a different loss
function, as demonstrated by~\cite{athalye2018obfuscated}.
In other words, gradient masking may make optimization-based attacks
fail but that does not mean that the space of adversarial perturbations decreased.

Defending against non-adaptive attacks is necessary but not sufficient.
It is our firm belief that \textbf{an evaluation against non-adaptive
  attacks is of very limited utility}.

Along the same lines, there is no justification to study a ``zero-knowledge''
\citep{biggio2013evasion} threat model where the attacker
is not aware of the defense.
``Defending'' against such an adversary is
an absolute bare-minimum that in no way suggests a defense will be effective
to further attacks. \cite{carlini2017adversarial} considered this scenario only to demonstrate
that some defenses were completely ineffective even against this very weak threat model.
The authors of that work now regret not making this explicit and
discourage future work from citing this paper in support of the zero-knowledge threat model.

It is crucial to actively attempt to defeat the specific defense being proposed.
On the most fundamental level
this should include a range of sufficiently different attacks with carefully tuned
hyperparameters.
But the analysis should go deeper than that:
ask why the defense might prevent existing attacks
from working optimally and how to customize existing
attacks or how to design completely new adversarial attacks to
perform as well as possible.
That is, applying the same mindset that a
future adversary would apply is the only way to show
that a defense might be able to withstand the test of time.

These arguments apply independent of the specific motivation of
the robustness evaluation: security, worst-case bounds
or human-machine gap all need a sense of the maximum vulnerability
of a given defense.
In all scenarios we should assume
the existence of an ``infinitely thorough'' adversary who will spend whatever time is
necessary to develop the optimal attack.

\subsection{Reproducible Research: Code \& Pre-trained Models}
\label{sec:releasecode}

Even the most carefully-performed robustness evaluations can have
subtle but fundamental flaws.
We strongly believe that releasing full source code and pre-trained models is one of the most
useful methods for ensuring the eventual correctness of an evaluation.
Releasing source code makes it much more likely that others will be able to
perform their own analysis of the defense.\footnote{In their analysis
  of the ICLR 2018 defenses \citep{athalye2018obfuscated}, the
  authors spent five times longer re-implementing the defenses than
  performing the security evaluation of the re-implementations.}
Furthermore, completely specifying all defense details in a paper can be
difficult, especially in the typical 8-page limit of many
conference papers.
The source code for a defense can be seen as the definitive
reference for the algorithm.

It is equally important to release pre-trained models, especially when
the resources that would be required to train a model would be prohibitive
to some researchers with limited compute resources.
The code and model that is released should be the model that was used
to perform the evaluation in the paper to the extent permitted by
underlying frameworks for accelerating numerical computations
performed in machine learning.
Releasing a \emph{different}
model than was used in the paper makes it significantly less useful,
as any comparisons against the paper may not be identical.

Finally, it is helpful if the released code contains a simple one-line
script which will run the full defense end-to-end on the given input.
Note that this is often different than what the defense developers want,
who often care most about performing the evaluation as efficiently as
possible.
In contrast, when getting started with evaluating a defense (or to
confirm any results), it is
often most useful to have a simple and correct method for running the
full defense over an input.

There are several frameworks such as CleverHans~\citep{papernot2018cleverhans} or Foolbox~\citep{rauber2017foolbox}
as well as websites\footnote{https://robust-ml.org}\textsuperscript{,}%
\footnote{https://robust.vision/benchmark/leaderboard/}\textsuperscript{,}%
\footnote{https://foolbox.readthedocs.io/en/latest/modules/zoo.html}
which have been developed to assist in this process.

\section{Specific Recommendations: Evaluation Checklist}
\label{sec:dont_do_bad_science}
\label{sec:pleaseactuallythink}
While the above overview is general-purpose advice we believe will stand the
test of time, it can be difficult to extract specific, actionable items
from it.
To help researchers \emph{today} perform more thorough evaluations,
we now develop a checklist that lists common evaluation
pitfalls when evaluating adversarial robustness. Items in this list are
sorted (roughly) into three categories.

The items contained below are \textbf{neither necessary nor sufficient}
for performing a complete adversarial example evaluation, and are
intended to list common evaluation flaws.
There likely exist completely ineffective defenses which satisfy all of
the below recommendations; conversely, some of the strongest defenses
known today do \emph{not} check off all the boxes below (e.g.\ \citet{madry2017towards}).

We encourage readers to be extremely careful and \textbf{not directly follow
  this list} to perform an evaluation or decide if an evaluation that has been
performed is sufficient.
Rather, this list contains common flaws that are worth checking for to
identify potential evaluation flaws.
Blindly following the checklist without careful thought will likely be counterproductive:
each item in the list must be taken into consideration within the context
of the specific defense being evaluated. 
Each item on the list below is present because we are aware of several defense
evaluations which were broken and following that specific recommendation would have
revealed the flaw.
We hope this list will be taken as a collection of recommendations that may
or may not apply to a particular defense, but have been useful in the past.

This checklist is a living document that lists the most common evaluation
flaws as of 18 February, 2019. We expect the evaluation flaws that are 
common today will \emph{not} be the most common flaws in the future.
We intend to keep this checklist up-to-date with the latest recommendations
for evaluating defenses by periodically updating its contents.
Readers should check the following URL  for the most recent
revision of the checklist:
\url{https://github.com/evaluating-adversarial-robustness/adv-eval-paper}.

\subsection{Common Severe Flaws}
There are several common severe evaluation flaws which have the
potential to completely invalidate any robustness claims.
Any evaluation which contains errors on any of the following
items is likely to have fundamental and irredeemable flaws.
Evaluations which intentionally deviate from the advice here may wish to
justify the decision to do so.

\begin{itemize}[leftmargin=*]
  \item \S\ref{sec:pleaseactuallythink} \textbf{Do not mindlessly follow this list}; make sure to still think about the evaluation.
\item \S\ref{sec:threatmodel} \textbf{State a precise threat model} that the defense is supposed to be effective under.
  \begin{itemize}[leftmargin=*]
  \item The threat model assumes the attacker knows how the defense works.
  \item The threat model states attacker's goals, knowledge and capabilities.
  \item For security-justified defenses, the threat model realistically models some adversary.
  \item For worst-case randomized defenses, the threat model captures the perturbation space.
  \item Think carefully and justify any $\ell_p$ bounds placed on the adversary.
  \end{itemize}
\item \S\ref{sec:adaptive} Perform \textbf{adaptive attacks} to give an upper bound of robustness.
  \begin{itemize}[leftmargin=*]
  \item The attacks are given access to the full defense, end-to-end.
  \item The loss function is changed as appropriate to cause misclassification.
  \item \S\ref{sec:whichattack} \textbf{Focus on the strongest attacks} for the threat model and defense considered.
  \end{itemize}
\item \S\ref{sec:releasecode} Release \textbf{pre-trained models and source code}.
  \begin{itemize}[leftmargin=*]
  \item Include a clear installation guide, including all dependencies.
  \item There is a one-line script which will classify an input example with the defense.
  \end{itemize}
\item \S\ref{sec:cleanaccuracy} Report \textbf{clean model accuracy} when not under attack.
  \begin{itemize}[leftmargin=*]
  \item For defenses that abstain or reject inputs, generate a ROC curve.
  \end{itemize}
\item \S\ref{sec:sanitycheck} Perform \textbf{basic sanity tests} on attack success rates.
  \begin{itemize}[leftmargin=*]
    \item Verify iterative attacks perform better than single-step attacks.
    \item Verify increasing the perturbation budget strictly increases attack success rate.
    \item With ``high'' distortion, model accuracy should reach levels of random guessing.
  \end{itemize}
\item \S\ref{sec:100success} Generate an \textbf{attack success rate vs. perturbation budget} curve.
  \begin{itemize}[leftmargin=*]
  \item Verify the x-axis extends so that attacks eventually reach 100\% success.
  \item For unbounded attacks, report distortion and not success rate.
  \end{itemize}
\item \S\ref{sec:whitebox} Verify \textbf{adaptive attacks} perform better than any other.
  \begin{itemize}[leftmargin=*]
  \item Compare success rate on a per-example basis, rather than averaged across the dataset.
  \item Evaluate against some combination of black-box, transfer, and random-noise attacks.
  \end{itemize}
\item \S\ref{sec:describeattacks} Describe the \textbf{attacks applied}, including all hyperparameters.

\end{itemize}

\subsection{Common Pitfalls}
There are other common pitfalls that may prevent the detection of ineffective defenses.
This list contains some potential pitfalls which do not apply to
large categories of defenses.
However, if applicable, the items below are still important to carefully
check they have been applied correctly.

\begin{itemize}[leftmargin=*]
\item \S\ref{sec:whichattack} Apply a \textbf{diverse set of attacks} (especially when training on one attack approach).
  \begin{itemize}[leftmargin=*]
  \item Do not blindly apply multiple (nearly-identical) attack approaches.
  \end{itemize}
\item \S\ref{sec:gradientfree} Try at least one \textbf{gradient-free attack} and one \textbf{hard-label attack}.
  \begin{itemize}[leftmargin=*]
  \item Try \cite{chen2017zoo,uesato2018adversarial,ilyas2018black,brendel2017decision}.
  \item Check that the gradient-free attacks succeed less often than gradient-based attacks.
  \item Carefully investigate attack hyperparameters that affect success rate.
  \end{itemize}
\item \S\ref{sec:transfer} Perform a \textbf{transferability attack} using a similar substitute model.
  \begin{itemize}[leftmargin=*]
    \item Select a substitute model as similar to the defended model as possible.
    \item Generate adversarial examples that are initially assigned high confidence.
    \item Check that the transfer attack succeeds less often than white-box attacks.
  \end{itemize}
\item \S\ref{sec:eot} For randomized defenses, properly \textbf{ensemble over randomness}.
  \begin{itemize}[leftmargin=*]
  \item Verify that attacks succeed if randomness is assigned to one fixed value.
  \item State any assumptions about adversary knowledge of randomness in the threat model.
  \end{itemize}
\item \S\ref{sec:bpda} For non-differentiable components, \textbf{apply differentiable techniques}.
  \begin{itemize}[leftmargin=*]
  \item Discuss why non-differentiable components were necessary.
  \item Verify attacks succeed on undefended model with those non-differentiable components.
  \item Consider applying BPDA~\citep{athalye2018obfuscated} if applicable.
  \end{itemize}
\item \S\ref{sec:converge} Verify that the \textbf{attacks have converged} under the selected hyperparameters.
  \begin{itemize}[leftmargin=*]
  \item Verify that doubling the number of iterations does not increase attack success rate.
  \item Plot attack effectiveness versus the number of iterations.
  \item Explore different choices of the step size or other attack hyperparameters.
  \end{itemize}
\item \S\ref{sec:hyperparams} Carefully \textbf{investigate attack hyperparameters} and report those selected.
  \begin{itemize}[leftmargin=*]
  \item Start search for adversarial examples at a random offset.
  \item Investigate if attack results are sensitive to any other hyperparameters.
  \end{itemize}
  \item \S\ref{sec:priorwork} \textbf{Compare against prior work} and explain important differences.
  \begin{itemize}[leftmargin=*]
  \item When contradicting prior work, clearly explain why differences occur.
  \item Attempt attacks that are similar to those that defeated previous similar defenses.
  \item When comparing against prior work, ensure it has not been broken.
  \end{itemize}
\item \S\ref{sec:generalrobustness} Test \textbf{broader threat models} when proposing general defenses. For images:
  \begin{itemize}[leftmargin=*]
  \item Apply rotations and translations \citep{engstrom2017rotation}.
  \item Apply common corruptions and perturbations \citep{hendrycks2018benchmarking}.
  \item Add Gaussian noise of increasingly large standard deviation \citep{ford2019adversarial}.
  \end{itemize}

\end{itemize}

\subsection{Special-Case Pitfalls}
The following items apply to a smaller fraction of evaluations.
Items presented here are included because while
they may diagnose flaws in
some defense evaluations, they are not necessary for many others.
In other cases, the tests presented here help provide additional evidence that the
evaluation was performed correctly.

\begin{itemize}[leftmargin=*]

\item \S\ref{sec:provable} Investigate if it is possible to use \textbf{provable approaches}.
  \begin{itemize}[leftmargin=*]
    \item Examine if the model is amenable to provable robustness lower-bounds.
  \end{itemize}
\item \S\ref{sec:randomnoise} \textbf{Attack with random noise} of the correct norm.
  \begin{itemize}[leftmargin=*]
    \item For each example, try 10,000+ different choices of random noise.
    \item Check that the random attacks succeed less-often than white-box attacks.
  \end{itemize}
\item \S\ref{sec:targeted} Use both \textbf{targeted and untargeted attacks} during evaluation.
  \begin{itemize}[leftmargin=*]
    \item State explicitly which attack type is being used.
  \end{itemize}
\item \S\ref{sec:attacksimilar} \textbf{Perform ablation studies} with combinations of defense components removed.
  \begin{itemize}[leftmargin=*]
  \item Attack a similar-but-undefended model and verify attacks succeed.
  \item If combining multiple defense techniques, argue why they combine usefully.
  \end{itemize}
\item \S\ref{sec:benchmarkattack} \textbf{Validate any new attacks} by attacking other defenses.
  \begin{itemize}[leftmargin=*]
  \item Attack other defenses known to be broken and verify the attack succeeds.
  \item Construct synthetic intentionally-broken models and verify the attack succeeds.
  \item Release source code for any new attacks implemented.
  \end{itemize}
\item \S\ref{sec:notimages} Investigate applying the defense to \textbf{domains other than images}.
  \begin{itemize}[leftmargin=*]
  \item State explicitly if the defense applies only to images (or another domain).
  \end{itemize}
  \item \mbox{\S\ref{sec:reportmeanmin} Report \textbf{per-example attack success rate}:
$\mathop{\text{mean}}\limits_{x \in \mathcal{X}} \min\limits_{a \in \mathcal{A}} f(a(x))$, not
$\mathop{\text{min}}\limits_{a \in \mathcal{A}} \mathop{\text{mean}}\limits_{x \in \mathcal{X}} f(a(x))$.}
\end{itemize}

\section{Evaluation Recommendations}

We now expand on the above checklist and provide the rationale for each item.

\subsection{Investigate Provable Approaches}
\label{sec:provable}

With the exception of this subsection, all other advice in this
paper focuses on performing heuristic robustness evaluations.
Provable robustness approaches are preferable to only heuristic
ones.
Current provable approaches often can only be applied when the neural
network is explicitly designed with the objective of making these specific
provable techniques applicable \citep{kolter2017provable,raghunathan2018certified,WengZCSHBDD18}.
While this approach of designing-for-provability has seen excellent
progress---the best approaches today can certify some (small) robustness
even on ImageNet classifiers \citep{lecuyer2018certified}---often the best
heuristic defenses offer orders of magnitude better (estimated) robustness.

Proving a lower bound of defense
robustness guarantees that the robustness will never fall
below that level (if the proof is correct).
We believe an important direction of future research is developing approaches
that can generally prove arbitrary neural networks correct.
While work
in this space does exist \citep{katz2017reluplex,TjengXT19,XiaoTSM19,GowalDSBQUAMK18}, it is often computationally
intractable to verify even modestly sized neural networks.

One key limitation of provable techniques is that the proofs they
offer are generally only of the form ``for some \emph{specific} set of
examples $\mathcal{X}$, no adversarial example with distortion less than
$\varepsilon$ exists''.
While this is definitely a useful statement, it
gives no proof about any \emph{other} example $x' \not\in \mathcal{X}$;
and because this is the property that we actually care about, provable
techniques are still not provably correct in the same way that provably
correct cryptographic algorithms are provably correct.

\subsection{Report Clean Model Accuracy}
\label{sec:cleanaccuracy}

A defense that significantly degrades the model's accuracy on the original
task (the \emph{clean} or \emph{natural} data) may not be useful
in many situations.
If the probability of an actual attack is very low and the cost of an 
error on adversarial inputs is not
high, then it may be unacceptable to incur \emph{any} decrease in clean
accuracy.
Often there can be a difference in the impact of an error on a random
input and an error on an adversarially chosen input. To what extent this
is the case depends on the domain the system is being used in.

For the class of defenses that refuse to classify inputs by abstaining when
detecting that inputs are adversarial, or otherwise refuse to classify
some inputs, it is important to evaluate how this impacts accuracy on
the clean data.
Further, in some settings it may be acceptable to refuse to classify inputs
that have significant amount of noise. In others, while it may be acceptable
to refuse to classify adversarial examples, simple noisy inputs must still
be classified correctly.
It can be helpful to generate a Receiver Operating Characteristic (ROC) curve
to show how the choice of threshold for rejecting inputs causes the clean
accuracy to decrease.

\subsection{Focus on the Strongest Attacks Possible}
\label{sec:whichattack}

\paragraph{Use optimization-based attacks.}
Of the many different attack algorithms, optimization-based attacks
are by far the most powerful.
After all, they extract a significant amount of information from the model
by utilizing the gradients of some loss function and not just the predicted output.
In a white-box setting, there are many different attacks
that have been created, and picking almost any of them will be useful.
However, it is important to \emph{not} just choose an attack and apply it out-of-the-box
without modification.
Rather, these attacks should serve as a starting point
to which defense-specific knowledge can be applied.

We have found the following three attacks useful starting points for constructing
adversarial examples under different distance metrics:
\begin{itemize}
  \item For $\ell_1$ distortion, start with \cite{chen2017ead}.
  \item For $\ell_2$ distortion, start with \cite{carlini2017towards}.
  \item For $\ell_\infty$ distortion, start with \cite{madry2017towards}.
\end{itemize}

Note, however, that these attacks were designed to be effective on standard
neural networks; any defense which modifies the architecture, training
process, or any other aspect of the machine learning algorithm is likely to affect their performance.
In order to ensure that they perform reliably on a particular defense,
a certain amount of critical thinking and hyper-parameter optimization is necessary.
Also, remember that there is
no universally \emph{strongest} attack: each attack makes specific assumptions
about the model and so an attack that is best on one model might perform
much worse on another \citep{schott2018}.
It is worth considering many different attacks.

\paragraph{Do not use Fast Gradient Sign (exclusively).}
The Fast Gradient Sign
(FGS) attack is a simple approach to generating adversarial examples
that was proposed to demonstrate the linearity of neural networks \citep{goodfellow2014explaining}.
It was never intended to be a strong attack for evaluating the robustness
of a neural network, and should not be used as one.

Even if it were intended to be a strong attack worth defending
against, there are many defenses which achieve near-perfect success at
defending against this attack.
For example, a weak version of
adversarial training can defend against this attack by causing
gradient masking  \citep{tramer2017ensemble},
where locally the gradient around a given image may
point in a direction that is not useful for generating an adversarial
example.

While evaluating against FGS can be a component of an attack
evaluation, it should never be used as the only attack in an
evaluation.

At the same time, even relatively simple and efficient attacks
(DeepFool and JSMA) \citep{moosavi2016deepfool,papernot2016limitations}
can still be useful for evaluating adversarial
robustness if applied correctly \citep{jetley2018friends,carlini2016defensive}.
However doing so often requires more care and is more error-prone;
applying gradient-based attacks for many iterations is often simpler
despite the attacks themselves being slower.

\paragraph{Do not \emph{only} use attacks during testing that were used during training.}
One pitfall that can arise with adversarial training is that the defense
can overfit against one particular attack
used during training (e.g. by masking the gradients)  \citep{tramer2017ensemble}.
Using the same attack during both training and testing is dangerous
and is likely to overestimate the robustness of the defense.
This is still true even if the attack is perceived as the
strongest attack for the given
threat model: this is probably not true any more given that the defense
has been specifically tuned against it.

\paragraph{Applying many nearly-identical attacks is not useful.}
When applying a diverse set of attacks, it is critical that the attacks
are actually diverse.
For example, the Basic Iterative Method (BIM) (also
called i-FGSM, iterated fast gradient sign) \citep{kurakin2016adversarial}
is nearly identical to Projected Gradient Descent \citep{madry2017towards}
modulo the initial random step.
Applying both of these attacks is less useful than applying one of these
attacks and another (different) attack approach.

\subsection{Apply Gradient-Free Attacks}
\label{sec:gradientfree}

To ensure that the model is not causing
various forms of gradient masking, it is worth attempting
gradient-free attacks.
There are several such proposals. The following
three require access to model confidence values, and therefore
some forms of gradient masking (e.g., by thresholding the model
outputs) will prevent these attacks from working effectively.
Nevertheless, these attacks are effective under many situations
where standard gradient-based attacks fail:
\begin{itemize}
  \item ZOO \citep{chen2017zoo} (and also \citep{LiuCLS17,bhagoji2018practical,TuTCLZYHC18}) numerically estimates gradients and then
performs gradient descent, making it powerful but potentially
ineffective when the loss surface is difficult to optimize over.
  \item SPSA \citep{uesato2018adversarial} was proposed
specifically to evaluate adversarial example defenses, and has broken
many. It can operate even when the loss surface is difficult to
optimize over, but has many hyperparameters that can be difficult to tune.
  \item NES \citep{ilyas2018black} is effective at generating adversarial examples with a
limited number of queries, and so generates adversarial examples of
higher distortion. One variant of NES can handle scenarios in which only the confidence
values of the top-$k$ classes or only the label corresponding to the most confident output (see below)
are observed.
This family of approaches can be further strengthened by including and transferring over the existing priors \citep{IlyasEM18}.
\end{itemize}

Hard label (or ``decision-based'') attacks differ from gradient-free confidence-based attacks
in that they only require access to the \texttt{arg max} output of the
model (i.e., the label corresponding to the most confident output).
This makes these attacks much slower, as they make many more queries,
but defenses can do less to accidentally prevent these attacks.

\begin{itemize}
  \item The Boundary Attack \citep{brendel2017decision} is general-purpose
  hard-label attack and performs a descent along the decision boundary using a rejection sampling approach.
  In terms of the minimum adversarial distance the attack often rivals the best white-box attacks 
  but at the cost of  many more queries.
\end{itemize}

The most important reason to apply gradient-free attacks is as a test
for gradient masking \citep{tramer2017ensemble,athalye2018obfuscated,uesato2018adversarial}.
Gradient-free attacks should almost always do worse than gradient-based
attacks (see \S\ref{sec:sanitycheck}).
Not only
should this be true when averaged across the entire dataset, but
also on a per-example basis (see \S\ref{sec:reportmeanmin}) there
should be very few instances where gradient-based attacks fail but
gradient-free attacks succeed.
Thus, white-box attacks performing
worse than gradient-free attacks is strong evidence for gradient masking.

Just as gradient masking can fool gradient-based attack algorithms,
gradient-free attack algorithms can be similarly fooled by different
styles of defenses.
Today, the most common class of models that causes existing
gradient-free attacks to fail is \emph{randomized models}.
This may change in the future as gradient-free attacks become stronger,
but the state-of-the-art gradient-free attacks currently do not do well in this setting.
(Future work on this challenging problem would be worthwhile.)

\subsection{Perform a Transferability Analysis}
\label{sec:transfer}

Adversarial examples often
transfer between models \citep{papernot2016transferability}.
That is, an adversarial example generated
against one model with a specific architecture will
often also be adversarial on another model, even if the latter is trained on a
different training set with a different architecture.

By utilizing this phenomenon, \emph{transfer attacks} can be performed by considering
an alternative, substitute model and generating adversarial examples for that are strongly
classified as a wrong class by that model.
These examples can then be used to attack the target model.
Such transfer attacks are particularly successful at circumventing
defenses that rely on gradient masking.
Since the adversarial example is generated on an independent model, there is no
need to directly optimize over the (often poorly behaved) landscape of the defended model.
An adversarial example defense evaluation should thus attempt a
transferability analysis to validate that the proposed
defense breaks transferability.
If it does not, then it is likely to
not actually be a strong defense, but just appears like one.

What should be the source model for a transferability analysis? A good
starting point is a model as similar to the defended model as
possible, trained on the same training data.
If the defended model
adds some new layers to a baseline model, it would be good to try the
baseline.
Using the undefended baseline allows optimization-based attacks to
reliably construct high-confidence adversarial examples that are likely to
transfer to the defended model.
We also recommend trying a strong adversarially
trained model \citep{madry2017towards}, which has been shown to be a strong source model for
adversarial examples.
It is also more effective to generate
adversarial examples that fool an ensemble of source models, and then
use those to transfer to a target model \citep{liu2016delving}.

\subsection{Properly Ensemble over Randomness}
\label{sec:eot}

For defenses that randomize aspects of neural network inference,
it is important to properly generate adversarial examples by
ensembling over the randomness of the defense.
The randomness introduced might make it difficult to apply standard attacks
since the output of the classifier as well the loss gradients used for
optimization-based attack now become stochastic.
It is often necessary to repeat each step of a standard attack multiple times
to obtain reliable estimates.
Thus, by considering multiple different choices of randomness, one can generate
adversarial examples that are still for a new choice of randomness.

Defenses should be careful that relying on an exponentially large
randomness spaces may not actually make attacks exponentially
more difficult to attack.
It is often the case that one can construct adversarial examples that are resistant
to such randomness by simply constructing examples that are consistently
adversarial over a moderate number of randomness choices.

\paragraph{Verify that attacks succeed if randomness is fixed.}
If randomness is believed to be an important reason why the defense
is effective, it can be useful to sample one value of randomness and
use only that one random choice.
If the attacks fail even when the randomness is disabled in this
way, it is likely that the attack is not working correctly and
should be fixed.
Once the attack succeeds with randomness
completely disabled, it can be helpful to slowly re-enable the
randomness and verify that attack success rate begins to decrease.

\subsection{Approximate Non-Differentiable Layers}
\label{sec:bpda}

Some defenses include non-differentiable layers as pre-processing
layers, or later modify internal layers to make them non-differentiable
(e.g., by performing quantization or adding extra randomness).
Doing this makes the defense much harder to evaluate completely:
gradient masking becomes much more likely when some layers are
non-differentiable, or have difficult-to-compute gradients.

In many cases, it can happen that one accidentally introduces non-differentiable
layers \citep{athalye2018obfuscated} (e.g., by introducing components
that, while differentiable, are not usefully-differentiable) or layers that
cause gradients to vanish to zero (e.g., by saturating
activation functions \citep{brendel2017comment,carlini2016defensive}).

In these cases it can be useful to create a
differentiable implementation of the layer (if possible) to
obtain gradient information.
In other case, applying BPDA \citep{athalye2018obfuscated} can help
with non-differentiable layers.
For example, if a defense implements a pre-processing layer that
denoises the input and has the property that $\text{denoise}(x) \approx x$,
it can be effective to approximate its gradient as the identity
function on the backward pass, but compute the function exactly
on the forward pass.

\subsection{Verify Attack Convergence}
\label{sec:converge}

On at least a few inputs, confirm
that the number of iterations of gradient descent being performed is
sufficient by plotting the attack success versus the number of
iterations.
This plot should eventually plateau; if not, the attack
has not converged and more iterations can be used until
it does.
The number of
iterations necessary is generally inversely proportional to step size and
proportional to distortion allowed.
Dataset complexity is often related.

For example, on
CIFAR-10 or ImageNet with a maximum $\ell_\infty$ distortion of 8/255,
white-box optimization attacks generally converge in under 100 or 1000
iterations with a step size of 1.
However, black-box attacks often take orders of magnitude more queries,
with attacks requiring over 100,000 queries not being abnormal.
For a different dataset, or a different
distortion metric, attacks will require a different number of iterations.
While it is
possible to perform an attack correctly with very few iterations
of gradient descent, it requires much more thought and care \citep{engstrom2018evaluating}.
For example, in some cases even a million iterations of white-box gradient descent
have been found necessary \citep{qian2018l2}.

There are few reasonable threat models under which an attacker can
compute 100 iterations of gradient descent, but not 1000.
Because the threat model must not constrain the approach an
attacker takes, it is worth evaluating against a strong attack
with many iterations of gradient descent.

\paragraph{Ensure doubling attack iterations does not increase attack success rate.}
Because choosing any fixed bounded number of iterations can be difficult to
generalize, one simple and useful method for determining if ``enough'' iterations
have been performed is to try doubling the number of iterations and check if this
improves on the adversarial examples generated.
This single test is not the only method to select the number of iterations,
but once some value has been selected, doubling the number of iterations
can be a useful test that the number selected is probably sufficient.

\subsection{Carefully Investigate Attack Hyperparmeters}
\label{sec:hyperparams}

\paragraph{Start search from random offsets.}
When performing iterative attacks,
it is useful to begin the search for adversarial examples at random
offsets away from the initial, source input.
This can reduce the
distortion required to generate adversarial examples by 10\%
\citep{carlini2017towards}, and help avoid causing gradient masking
\citep{tramer2017ensemble,madry2017towards}.
Doing this
every time is not necessary, but doing it at least once to verify it
does not significantly change results is worthwhile.
Prior work has found that using at least 10 random starting points is effective.

\paragraph{Carefully select attack hyperparameters.}
Many attacks support a wide range of hyperparameters.
Different settings of these parameters
can make order-of-magnitude differences in attack success rates.
Fortunately, most hyperparameters have the property that moving them in
one direction strictly increases attack power (e.g., more iterations of
gradient descent is typically better).
When in doubt, chose a stronger set of hyperparameters.

All hyperparameters and the way in which they were determined should be reported.
This allows others to reproduce the results and helps to
understand how much the attacks
have been adapted to the defense.

\subsection{Test General Robustness for General-Purpose Defenses}
\label{sec:generalrobustness}
Some defenses are designed not to be robust against any one specific
type of attack, but to generally \emph{be robust}.
This is in contrast to other defenses, for example adversarial training
\citep{goodfellow2014explaining,madry2017towards}, which explicitly set out
to be robust against one specific threat model (e.g., $l_\infty$ adversarial
robustness).

When a defense is arguing that it generally improves robustness, it can
help to also verify this fact with easy-to-apply attacks.
Work in this space exists mainly on images; below we give three
good places to start, although they are by no means the only
that are worth applying.
\begin{itemize}
\item Transform the image with a random rotation and translation
  \citep{engstrom2017rotation}. This attack can be performed
  brute-force and is thus not susceptible to gradient masking.
\item Apply common corruptions and perturbations
  \citep{hendrycks2018benchmarking} that mimic changes that may actually
  occur in practice.
\item Add Gaussian noise with increasingly large standard deviation
  \citep{ford2019adversarial}. Adversarially robust models tend to be more resistant to random noise compared to their standard counterparts
  \citep{fawzi2016robustness}.
\end{itemize}
In all cases, if these tests are used as a method for evaluating general-purpose
robustness it is important to \emph{not train on them directly}: doing so
would counter their intended purpose.

\subsection{Try Brute-Force (Random) Search Attacks}
\label{sec:randomnoise}

A very simple sanity check to ensure that the attacks have not
been fooled by the defense is trying random search to
generate adversarial examples within the threat model.
If brute-force random sampling
identifies adversarial examples that other methods haven't found, this
indicates that other attacks can be improved.

We recommend starting by sampling random points at a large distance from the original input.
Every time
an adversarial example is found, limit the search to adversarial
examples of strictly smaller distortion.
We recommend verifying a
hundred instances or so with a few hundred thousand random samples.

\subsection{Targeted and Untargeted Attacks}
\label{sec:targeted}
In theory, an untargeted attack is strictly easier than a targeted attack.
However, in practice, there can be cases where targeting any of the $N-1$ classes
will give superior results to performing one untargeted attack.

At the implementation level, many untargeted attacks work by \emph{reducing}
the confidence in the correct prediction, while targeted attacks work by
\emph{increasing} the confidence in some other prediction.
Because these two formulations are not directly inverses of each other,
trying both can be helpful.

\subsection{Attack Similar-but-Undefended Models}
\label{sec:attacksimilar}

Defenses typically are implemented by making a series of changes
to a base model.
Some changes are introduced in order to increase
the robustness, but typically other changes are also introduced to
counteract some unintended consequence of adding the defense components.
In these cases, it can be useful to remove all of the defense
components from the defended model and attack a model with only the
added non-defense components remaining.
If the model still appears robust with these components not intended
to provide security, it is likely the attack is being artificially
broken by those non-defense components.

Similarly, if the defense has some tunable constants where changing
(e.g., by increasing)
the constant is believed to make generating adversarial examples
harder, it is important to show that when the constant is not
correctly set (e.g., by decreasing it) the model is vulnerable to
attack.

\subsection{Validate any new attack algorithms introduced}
\label{sec:benchmarkattack}

Often times it can be necessary to introduce a new attack algorithm
tailored to some specific defense.
When doing so, it is important
to carefully evaluate the effectiveness of this new attack algorithm.
It is unfortunately
easy to design an attack that will never be effective, regardless
of the model it is attacking.

Therefore, when designing a new attack, it is important to validate
that it is indeed effective.
This can be done by selecting alternate
models that are either known to be insecure or are intentionally
designed to be insecure, and verify that the new attack algorithm
can effectively break these defenses.

\section{Analysis Recommendations}
\label{sec:analysis}

After having performed an evaluation (consisting of at least 
some of the above
steps), there are several simple checks that will help
identify potential flaws in the attacks that should be corrected.

\subsection{Compare Against Prior Work}
\label{sec:priorwork}

Given the extremely large quantity of defense work in the space of
adversarial examples, it is highly unlikely that any idea is completely
new and unrelated to any prior defense.
Although it can be time-consuming, it is important to review prior
work and look for approaches that are similar to the new defense
being proposed.

This is especially important in security because any attacks which
were effective on prior similar defenses are likely to still
be effective.
It is therefore even more important to review not
only the prior defense work, but also the prior attack work to ensure
that all known attack approaches have been considered.
An unfortunately large number of defenses have been defeated by
applying existing attacks unmodified.

\paragraph{Compare against true results.}
When comparing against prior
work, it is important to report the accuracy of prior defenses under
the strongest attacks on these models.%
If a defense claimed that its
accuracy was $99\%$ but followup work reduced its accuracy to $80\%$,
future work should report the accuracy at $80\%$, and \textbf{not} $99\%$.
The original result is wrong.

There are many examples of defenses building on prior work which
has since been shown to be completely broken; performing a literature
review can help avoid situations of this nature. Websites such
as \texttt{RobustML}%
\footnote{https://www.robust-ml.org/},
are explicitly designed to help track defense progress.

\subsection{Perform Basic Sanity Tests}
\label{sec:sanitycheck}

\paragraph{Verify iterative attacks perform better than single-step attacks.}
Iterative attacks are strictly more powerful than single-step
attacks, and so their results should be strictly superior.
If an
iterative attack performs worse than a single-step attack, this often
indicates that the iterative attack is not correct.

If this is the case, one useful diagnostic test is to plot attack
success rate versus number of attack iterations averaged over many
attempts, to try and identify if there is a pattern.
Another diagnostic is to plot the model loss versus the number of attack
iterations for a single input.
The model loss should be (mostly) decreasing at each iteration.
If this is not the case, a smaller step size may be helpful.

\paragraph{Verify increasing the perturbation budget strictly
increases attack success rate.}
\label{sec:monodistortion}
Attacks that allow more distortion are strictly stronger
than attacks that allow less distortion.
If the attack success rate
ever decreases as the amount of distortion allowed increases, the
attack is likely flawed.

\paragraph{With ``high'' distortion, model accuracy should reach
  levels of random guessing.}
Exactly what ``high'' means will depend on the dataset.
On some datasets (like MNIST) even with noise bounded by an
$\ell_\infty$ norm of $0.3$, humans can often determine what
the digit was.
However, on CIFAR-10, noise with an $\ell_\infty$ norm of $0.2$ makes most
images completely unrecognizable.

Regardless of the dataset, there are some accuracy-vs-distortion
numbers that are theoretically impossible.
For example, it is not
possible to do better than random guessing with a $\ell_\infty$
distortion of $0.5$: any image can be converted into a solid gray
picture.
Or, for MNIST, the median $L_2$ distortion between a
digit and the most similar other image with a different label
is less than $9$, so claiming higher than this is not feasible.

\subsection{Generate An Accuracy versus Perturbation Curve}
\label{sec:100success}

One of the most useful diagnostic curves to generate is an accuracy versus perturbation budget
curve, along with an attack success rate versus perturbation budget curve.
These curves can help perform many of the sanity tests discussed in \S\ref{sec:sanitycheck}.

For some attack which produces minimally-distorted adversarial
examples~\citep{carlini2017towards} (as opposed to maximizing loss
under some norm constraint \citep{madry2017towards})
generating these curves is computationally efficient
(adversarial examples can be sorted by distance and then generating these curve
can be accomplished by sweeping a threshold constant).
For attacks that maximize loss given a fixed budget \citep{madry2017towards},
generating this curve naively requires calling the attack once for each value
of the perturbation budget, but this can be made more efficient by performing
binary search on the budget on a per-example basis.

\paragraph{Perform an unbounded attack.}
With unbounded
distortion, any attack should eventually reach 100\% success, even if only
by switching the input to actually be an instance of the other class.
If unbounded attacks do not succeed, this indicates that the attack is
being applied incorrectly.
The curve generated should have an x-axis with sufficient perturbation
so that it reaches 100\% attack success (0\% model accuracy).

\paragraph{For unbounded attacks, measure distortion, not success rate.}
The correct metric for evaluating unbounded attacks is the distortion
required to generate an adversarial example, not the success rate
(which should always be 100\%).
The most useful plot is to show success rate (or model accuracy) vs. distortion,
and the most useful single number is
either the mean or median distance to adversarial examples, when using
unbounded attacks.
To make measuring success rate meaningful, another
option is to artificially bound the attack and report any adversarial
example of distance greater than some threshold a failure (as long as
this threshold is stated).

\subsection{Verify white-box attacks perform better than black-box attacks}
\label{sec:whitebox}
Because white-box attacks are a strict super-set of black-box
attacks, they should perform strictly better.
In particular, this implies that gradient-based attacks should, in principle, outperform gradient-free attacks.
Gradient-free attacks doing
better often indicates that the defense is somehow masking gradient
information and that the gradient-based attack could be
improved to succeed more often.

When this happens, look at instances for which adversarial
examples can be found with black-box but not white-box attacks.
Check if these instances are at all related, and investigate why white-box
attacks are not finding these adversarial examples.

\subsection{Investigate Domains Other Than Images}
\label{sec:notimages}

Adversarial examples are not just a problem for image classification,
they are also a problem on sentiment analysis, translation,
generative models, reinforcement learning, audio classification,
and segmentation analysis (among others).
If a defense is limited to one domain, it should state this
fact explicitly.
If not, then it would be useful to briefly consider at least one
non-image domain to investigate if the technique could apply
to other domains as well.
Audio has properties most similar to images (high dimensional
mostly-continuous input space) and is therefore an easy point
of comparison.
Language processing is much different (due to the
inherently discrete input space) and therefore defenses are
often harder to apply to this alternate domain.

\subsection{Report the Per-Example Attack Success Rate}
\label{sec:reportmeanmin}

When evaluating attacks, it is important to report attack success rate
on a per-example
basis instead of averaged over the entire attack.
That is, report
$\mathop{\text{mean}}\limits_{x \in \mathcal{X}} \min\limits_{a \in \mathcal{A}} f(a(x))$, not
$\mathop{\text{min}}\limits_{a \in \mathcal{A}} \mathop{\text{mean}}\limits_{x \in \mathcal{X}} f(a(x))$.

This per-example reporting is strictly more useful than a per-attack reporting,
and is therefore preferable.
This is true despite the fact that in practice the results for the \emph{worst} attack
is often very close to the true per-example worst-case bounds.

This reporting also avoids another common pitfall in which two defenses A and B are compared
on different attacks X and Y leading to statements such as ``A is more robust than B against attack X
while B is more robust than A on attack Y''.
Such results are not useful, even more so if one attack
is strong and the other is weak.
Instead, by comparing defenses on
a per-example based (i.e. the optima per example over all attacks),
one defense can be selected as being stronger.
(Note this still requires the defenses are evaluated \emph{under the same threat model}.
It is not meaningful to compare a $\ell_\infty$-robust defense against a $\ell_0$-robust
defense.)

\subsection{Report Attacks Applied}
\label{sec:describeattacks}
After having performed a complete evaluation, as discussed above, defense
evaluations should report all attacks and
include the relevant hyperparameters.
It is especially important to report the number of iterations for optimization-based
attacks.

Note that it is not necessary to report the results from every
single attack in all details, especially if 
an attack is highly similar to another and yields similar results.
In this case the space can
rather be used to describe how the attacks were applied.
A good guideline is as follows: do
not show a massive table with many similar attacks that
have not been adapted to the defense.
Instead, rather show a shorter table with a more
diverse set of attacks that have each been
carefully adapted and tuned.

\section{Conclusion}

Evaluating adversarial example defenses requires extreme caution
and skepticism of all results obtained.
Researchers must be very careful
to not deceive themselves unintentionally when performing
evaluations.

This paper lays out the motivation for how to perform defense evaluations
and why we believe this is the case.
We develop a collection of recommendations
that we have identified as common flaws in adversarial
example defense evaluations.
We hope that this checklist will be
useful both for researchers developing novel defense approaches as
well as for readers and reviewers to understand if an evaluation is thorough
and follows currently accepted best practices.

We do not intend for this paper to be the definitive
answer to the question ``\emph{what
experiments should an evaluation contain?}''. Such a comprehensive list would be
impossible to develop.
Rather, we hope that
the recommendations we offer here will help inspire researchers
with ideas for how to perform their own evaluations.
\footnote{We believe that even if all defenses followed all of the
tests recommended in this paper, it would still be valuable
for researchers to perform re-evaluations of proposed defenses
(for example, by following the advice given in this paper and investigating
new, adaptive attacks).
For every ten defense papers on arXiv, there is just one paper which sets
out to re-evaluate previously proposed defenses.
Given the difficulty of performing
evaluations, it is always useful for future work to perform additional
experiments to validate or refute the claims made in existing papers.}

In total, we firmly believe that developing robust machine learning models is
of great significance, and hope that this document will in some way help
the broader community reach this important goal.

\section*{Acknowledgements}
We would like to thank Catherine Olsson and {\'U}lfar Erlingsson
for feedback on an early draft of this paper, and David Wagner for
helpful discussions around content.

\bibliography{paper}
\bibliographystyle{iclr2019_conference}

\end{document}